\documentclass[final,5p,times,twocolumn]{elsarticle}

\usepackage{amsmath}
\usepackage{amsfonts}
\usepackage{amssymb}
\usepackage{newtxtext,newtxmath}
\usepackage{bm}

\usepackage{array}
\usepackage{booktabs}
\usepackage{multirow}
\usepackage{multicol}
\newcolumntype{M}[1]{>{\centering\arraybackslash}m{#1}}
\newcommand{\tabincell}[2]{\begin{tabular}{@{}#1@{}}#2\end{tabular}}

\usepackage{graphicx}
\graphicspath{{figures/}}
\usepackage{stfloats}
\usepackage{float}
\usepackage{placeins}
\usepackage[ruled,vlined,linesnumbered]{algorithm2e}
\SetKwInput{KwInput}{Input}
\SetKwInput{KwOutput}{Output}
\SetKwProg{Fn}{Procedure}{}{}
\DontPrintSemicolon

\usepackage{siunitx}
\usepackage{nomencl}
\makenomenclature
\usepackage{comment}
\usepackage{soul}
\usepackage[normalem]{ulem}
\useunder{\uline}{\ul}{}

\usepackage{xcolor}

\usepackage{lineno}
\modulolinenumbers[5]
\usepackage[hidelinks]{hyperref}
\usepackage{xurl}

\journal{Next Energy}

\begin{document}

\begin{frontmatter}

\title{Uncovering Residential PV--EV Co-Adoption from Smart-Meter Data: Load Archetypes and Detection for Demand-Side Planning}

\author[label1]{Jack Zheng}
\ead{jack.zheng@monash.edu}
\author[label1,label2]{Hao Wang\corref{cor1}}
\ead{hao.wang2@monash.edu}
\cortext[cor1]{Corresponding author: Hao Wang.}
\affiliation[label1]{
            organization={Department of Data Science and AI, Faculty of Information Technology, Monash University},
            city={Melbourne},
            state={Victoria},
            country={Australia}
}
\affiliation[label2]{
            organization={Monash Energy Institute, Monash University},
            city={Melbourne},
            state={Victoria},
            country={Australia}
}

\begin{abstract}
The increasing adoption of electric vehicles (EVs) and rooftop photovoltaic (PV) systems is reshaping residential electricity demand and creating new challenges for demand-side management (DSM), tariff design, and low-voltage network planning. Much of the existing literature examines EV charging or PV generation in isolation, leaving the behavioral dynamics of household co-adoption less understood. We develop an integrated, two-part workflow to analyze advanced metering infrastructure (AMI) data. A discovery component applies dynamic time warping (DTW) $k$-means with DTW barycenter averaging to cluster daily import or export profiles into interpretable behavioral archetypes, while a predictive component trains a bidirectional long short-term memory (BiLSTM) model on 21-day windows and benchmarks it against tabular baselines for PV/EV activity detection. The EV activity labels are inferred from charging-like load signatures because charger measurements are unavailable. Using half-hourly AusNet residential data from Victoria, Australia, the clustering uncovers distinct patterns across PV-only, EV-only, co-adoption, and neither cohorts; for co-adopters, a midday-centered weekday export archetype accounts for approximately 50\% of days. At validation-tuned thresholds, both BiLSTM and XGBoost achieve strong discrimination. BiLSTM obtains $0.991$ for the area under the receiver operating characteristic curve (AUROC), $0.906$ for macro-F1, and the highest recall on the most difficult class ($0.836$ for EV-only recall). Tree-based baselines remain competitive. Performance remains stable across plausible labeling rules (macro-F1: $0.894$--$0.914$) and strictly forward temporal splits (macro-F1: $0.894$--$0.906$). The results show that shape-aware clustering and window-level sequence classification can provide a scalable AMI-based workflow for characterizing PV/EV-related load-shape behaviors and detecting activity patterns relevant to DSM targeting, tariff design, and low-voltage network planning.
\end{abstract}

\begin{keyword}
Electric vehicles \sep Photovoltaics \sep Residential load profiling \sep Smart-meter data \sep Non-intrusive load monitoring \sep Demand-side management


\end{keyword}
\end{frontmatter}

\section{Introduction} \label{sec:intro}
The increasing uptake of distributed energy resources (DERs), in particular electric vehicle (EV) adoption \cite{iea2024ev} and rooftop photovoltaic (PV) installations \cite{iea_solar}, is reshaping residential electricity demand profiles and energy planning requirements. Surveys of smart-meter analytics demonstrate that household interval data support descriptive, predictive, and prescriptive applications, including load profiling, forecasting, and demand-side interventions. These data also provide a methodological foundation for large-scale, non-intrusive analysis of end-use behaviors from aggregate traces \cite{wang2024ai,bai2026machine}. In addition to general analytics, shape-aware segmentation of daily profiles has become increasingly prevalent, with recent studies explicitly modeling weekday and weekend differences and recurring lifestyle-related patterns using real-world data \cite{tang2022}.

Empirical work has consistently found that home charging clusters in the late afternoon and evening, contributing to feeder stress \cite{neaimeh2015}, whereas rooftop PV depresses net load around midday and early afternoon. Together, these patterns motivate shifting home EV charging to daylight hours to ease existing peaks and make better use of renewable output \cite{sorensen2021}. Studies of co-adoption further show that households with both EVs and PV often move charging times into periods with solar availability, lowering evening demand and raising PV self-consumption \cite{liang2022}. At the program level, demand-side management (DSM) and price-responsive strategies can materially shift charging away from peak periods and towards renewable availability \cite{finn2012,denholm2013}. However, poorly designed synchronized responses to time-of-use tariffs can create new peaks instead of relieving the peak-time grid stress that drives costly infrastructure upgrades \cite{jones2022,needell2023}.

Utilities and network planners require scalable methods that (a) characterize how residential demand varies by technology ownership and household routines, and (b) non-intrusively determine which households or windows show PV/EV activity using advanced metering infrastructure (AMI) data already collected by utilities. Shape-aware clustering of daily profiles serves as a practical first step because it can reveal weekday and weekend archetypes and timing differences, such as solar-aligned troughs and evening-peak use, that bear directly on feeder loading and program design \cite{tang2022,hu2021}. Supervised detection of EV and PV activity from aggregate net-load traces can then help utilities segment customers for tariff offers or targeted plans without intrusive sub-metering. EV charging events can be detected from smart-meter recordings in both offline and online settings \cite{martin2023non}, while behind-the-meter PV generation can be sized or disaggregated from net load \cite{moreno2021,mason2020}; joint approaches recover several distributed energy resources from the same net-demand signal \cite{moreno2022}.

This ordering also matches common DSM and tariff objectives. Interpretable load-profile clusters allow planners to determine when peaks occur and when shifting may be feasible. Grouping households with similar behavioral patterns also enables interventions at scale. Previous studies show that managed or price-responsive charging can reduce evening peaks and better align demand with renewable production \cite{needell2023}, although poorly coordinated responses can cause new peaks to form \cite{jones2022}. Targeting therefore needs to separate cohorts whose charging is already solar-aligned from those whose evening charging would compound peaks.

Utilities are particularly concerned with load profiles that repeatedly coincide with constrained periods, including evening peaks, midday reverse-flow periods, and steep ramping intervals. The central research question in this paper is therefore whether residential AMI data can reveal distinct load-shape archetypes among PV-only, EV-only, PV--EV co-adopting, and non-adopting households, and how reliably PV/EV-associated activity windows can be detected from aggregate net-load traces to support demand-side planning.

This study combines daily-profile clustering and window-level classification in a single AMI workflow. The first stage characterizes recurring weekday and weekend load shapes across the four adoption cohorts; the second detects PV/EV-associated activity in 21-day household windows. The study makes four contributions.
\begin{itemize}
  \item  We characterize weekday and weekend daily import/export archetypes across the four PV/EV adoption states using dynamic time warping (DTW) $k$-means with DTW barycenter averaging (DBA) centroids, providing an interpretable view of recurring grid-exchange patterns by technology ownership.

  \item  We propose a multi-label detection approach that maps 21-day import windows to joint PV and EV activity labels through a bidirectional long short-term memory (BiLSTM) model with independently validation-tuned per-label thresholds. The sequence model is competitive with strong tabular baselines and provides higher recall on the most difficult (EV-only) class, while XGBoost achieves the strongest macro-F1.

  \item  We quantify the sensitivity of detection performance to plausible activity-labeling rules and to strictly forward temporal splits, separating the claims that are robust from those that depend on label construction.

  \item  We translate the behavioral and detection results into practical implications for DSM targeting, tariff design, and low-voltage network planning.
\end{itemize}

The remainder of this paper is organized as follows: Section~\ref{sec:literature-review} reviews the relevant literature. Sections~\ref{sec:study-design} and~\ref{sec:method} describe the study design and methods, respectively. Section~\ref{sec:results} presents the results, followed by robustness analyses in Section~\ref{sec:robustness} and discussion in Section~\ref{sec:discussion}. Section~\ref{sec:conclusion} concludes the paper.

\section{Literature Review}\label{sec:literature-review}
The literature reviewed here covers three main ideas: clustering of residential load profiles to detect and interpret common behaviors in households with PV, EV, or both; non-intrusive detection of PV/EV from aggregate smart-meter data; and translation of these outputs into utility-facing insights for DSM, tariff design, and low-voltage network planning strategies. Table~\ref{tab:litreview} groups the reviewed studies along these strands and indicates how each relates to the present work; the remainder of this section discusses them in turn.

\begin{table*}[tbp]
\centering
\caption{Summary of the reviewed literature by research theme, with representative studies and their relation to the present work.}
\label{tab:litreview}
\small
\setlength{\tabcolsep}{6pt}
\renewcommand{\arraystretch}{1.25}
\begin{tabular}{@{}p{3.2cm} p{3.5cm} p{9.2cm}@{}}
\toprule
\textbf{Research theme} & \textbf{Representative studies} & \textbf{Focus and relevance to this study} \\
\midrule
Residential load-shape clustering & \cite{tang2022,hu2021,zhang2025} & Shape- and feature-based segmentation of daily profiles, including EV-user clustering, supporting the descriptive archetype analysis. \\
Behind-the-meter PV inference & \cite{mason2020,moreno2021,moreno2022} & Estimate PV size and orientation and disaggregate PV and other DERs from net load, providing baselines for the PV side of detection. \\
EV detection from AMI & \cite{martin2023non,xiang2021,li2024,nandkeolyar2022} & Supervised, event-based, and training-free EV detection for demand-response targeting, providing baselines for the EV side of detection. \\
EV charging behavior and flexibility & \cite{neaimeh2015,sorensen2021,ziras2024,pan2019} & Characterize charging timing, flexibility, and distribution-network impacts that inform the activity-labeling rules. \\
PV--EV co-adoption & \cite{liang2022,bonin2022,sharda2024,bull2025} & Document behavioral change, incentives, and barriers under joint ownership, providing the empirical basis for interpreting co-adoption archetypes. \\
DSM, time-of-use tariffs, and managed charging & \cite{finn2012,denholm2013,jones2022,needell2023,liang2020,yang2024,desa2023} & Quantify peak-shifting benefits, solar-aligned charging, and synchronization risks that motivate cohort-aware targeting. \\
Integrated residential PV--EV(--storage) operation & \cite{fachrizal2021,saleem2024,nishanthy2023,fu2025} & Hosting-capacity, energy-management, and field studies of coordinated PV--EV operation that situate the downstream planning use cases. \\
Battery capacity and state estimation & \cite{wang2025hlhgur,wang2025psupf,wang2026socreview} & SOC/SOE estimation, temperature-aware filtering, and method comparisons relevant to coordinated PV--EV--storage operation. \\
\bottomrule
\end{tabular}
\end{table*}

Broad surveys cover the uses and potential of smart-meter analytics, with emphasis on workflows that pair discovery with prediction for utility use cases \cite{yildiz2017,bai2026machine}. This pairing sets the high-level design used here: first understand heterogeneity in load shapes and household behaviors, then build a classification model for detection at scale.

Residential profiling studies conducted over the last decade consistently report that clustering can uncover lifestyle-related patterns. Weekday and weekend profiles also exhibit behavioral differences \cite{yildiz2017}. Recent work combining interval demand with household context further shows that shape-aware clustering can help discover behavioral differences that are interpretable and policy-relevant, such as daytime versus evening peak usage \cite{tang2022}. Prior feature-based work also shows that PV and non-PV homes can be separated using intraday characteristics without intrusive sub-metering \cite{hu2021}. Profile clustering is therefore a well-supported discovery step for mapping behaviors across technology types.

Beyond feature-based methods, other studies detect DER signatures directly from aggregate smart-meter traces. For PV, deep learning and machine learning techniques have been used to infer installation parameters or disaggregate PV generation from net load \cite{mason2020,moreno2021}. For EV analysis, the literature includes supervised approaches \cite{martin2023non}, training-free optimization to detect charging events \cite{li2024}, and event-based extraction that uses the temporal structure of AMI data \cite{xiang2021}. Tabular models trained on window summaries have also been shown to detect DER presence effectively while keeping model complexity and computational cost low \cite{moreno2022}. This body of work justifies a supervised classifier for joint PV/EV activity detection and supplies the baselines against which the sequence model is compared.

Empirical evidence from smart-meter and charger datasets shows that at-home EV charging often concentrates from late afternoon to early evening, while PV production tends to depress midday net loads \cite{neaimeh2015}. Studies of apartments and other residential dwellings further explore charging flexibility and plug-in durations relevant to load shifting \cite{sorensen2021}. DSM optimization and system simulations show that managed or price-responsive charging can shave peaks and improve the use of renewable energy \cite{finn2012,denholm2013}. However, poorly coordinated responses to uniform time-of-use signals can create synchronized peaks instead \cite{jones2022,needell2023}. Because the same tariff can shave or synchronize peaks depending on which households respond, these risks support cohort-aware targeting rather than uniform signals.

Co-adoption studies using household-level smart-meter data report that households with both EV and PV tend to adjust charging habits towards daytime solar peaks under DSM-targeted incentives, lowering evening imports and promoting PV self-consumption \cite{liang2022}. This shift gives an empirical anchor for interpreting midday-centered clusters and for targeting solar-aligned behavior in practice.

Battery storage can also support coordinated PV--EV operation. Its contribution depends on available capacity and on reliable estimates of state of charge (SOC) and state of energy (SOE). Recent studies develop a history-gated recurrent model for online SOC/SOE co-estimation and a temperature-aware particle-filtering method for energy-state estimation under changing operating conditions \cite{wang2025hlhgur,wang2025psupf}. A recent review compares model-based, data-driven, and hybrid state-estimation methods in terms of accuracy, robustness, complexity, and measurement requirements \cite{wang2026socreview}. These battery-level methods are relevant when household AMI planning includes storage or controlled charging.

Prior work provides strong foundations in shape-aware clustering of daily load shapes \cite{tang2022,hu2021}, PV inference from net loads \cite{mason2020,moreno2021}, and EV detection through supervised or event-based logic \cite{martin2023non,li2024,xiang2021}. Less common is a single empirical AMI workflow that yields interpretable weekday and weekend archetypes across PV-only, EV-only, co-adoption, and neither cohorts and also evaluates joint PV/EV activity detection under household-level splits, validation-tuned thresholds, label-sensitivity checks, and forward temporal splits. Existing studies either document effects without a deployable detector \cite{liang2022} or detect a single technology in isolation, which limits DSM and tariff decisions where co-adoption influences behavior and joint detection matters \cite{jones2022,needell2023}.

\section{Data, Labels, and Study Design}\label{sec:study-design}

We use de-identified residential AMI data from AusNet in Victoria, Australia, recorded at 30-minute resolution (48 intervals per day), with separate grid-import and, where available, PV-export streams \cite{li2025wp12}. PV association is defined by an observable export stream and may therefore omit export-limited or zero-export PV systems. EV-associated households are identified using a rebate-status field in the supplied metadata, which is set to ``Payment Processed'' for EV-rebate recipients. For defining EV ownership, the rebate-status field is a reliable proxy, as the rebate provided a strong incentive for EV adoption among households that were early adopters in Australia. Standard data-cleaning procedures remove records with incomplete location information and duplicate daily entries.

The 30-minute sampling interval is typical of utility AMI deployments and is well suited to daily load-shape analysis, although it limits the granularity with which EV charging can be resolved. At this resolution, short charging events, partial top-ups, and appliance cycles with similar step changes may be compressed into the same interval. Consequently, the detection task is formulated at the window level: the classifier determines whether a 21-day window is PV-positive or EV-positive under the specified rules, without attempting to localize every individual charging interval.

Based on the PV/EV indicators, households are grouped into four non-overlapping cohorts: neither, PV-only, EV-only, and co-adoption. The clustering analysis uses data from 1 April 2022 to 31 October 2022, preserving weekday, weekend, and seasonal context while maintaining manageable computational requirements. The classification analysis uses data from 1 January 2020 to 30 April 2023, providing more diverse conditions for evaluating detection performance and temporal generalization.

The two date ranges are selected for distinct purposes. The clustering period is deliberately narrower because of the computational intensity of DTW clustering and the objective of recovering interpretable, recurring daily shapes, for which exhaustive time coverage is unnecessary. The classification period is broader to ensure that the detection model encounters a wider range of weather, occupancy, and seasonal PV conditions. This separation maintains the analytical separation of the clustering and classification stages. Both analyses use the same underlying data source but address different research objectives: the former is descriptive and the latter predictive.

For descriptive analysis, daily profiles $x \in \mathbb{R}^{48}$ are grouped by cohort and day type (weekday or weekend) to detect recurring load-shape archetypes. For predictive analysis, each household's import series is segmented into sliding windows of 21 consecutive days with a 7-day stride and mapped to two binary target labels, $y_{\mathrm{PV}}$ and $y_{\mathrm{EV}}$. The corresponding pair of model predictions maps to the four household-window classes used in the model comparison.

The use of 21-day windows reflects a compromise between behavioral coverage and model tractability. A window of three weeks captures repeated routines and multiple opportunities for EV charging while remaining short enough to keep sequences tractable. The 7-day stride preserves weekly alignment between adjacent windows and reduces sensitivity to an arbitrary window start date. Because adjacent windows from the same household can overlap, household-level splitting is essential; otherwise, nearly identical behavioral sequences could appear in both training and testing sets.

Day-level activity rules are used to construct window labels. A PV-active day is defined by the presence of a PV-export stream for that household, with a stricter sensitivity variant requiring same-day PV export $>0$\,kWh. Because charger-level measurements are unavailable, EV-active days are inferred from charging-like increases in aggregate smart-meter import data relative to fixed within-day reference blocks. For household $h$, day $d$, and half-hour bin $t$, let $x_{hdt}$ denote the metered half-hour energy and $\bar{x}_{hd,A}$ its mean over block $A$. The implementation forms
\begin{equation}
\begin{aligned}
c^{\mathrm{eve}}_{hdt} &= x_{hdt}-\bar{x}_{hd,16{:}00\text{--}19{:}30},
&t&\in \substack{20{:}00\text{--}23{:}30\\\text{or }00{:}00\text{--}01{:}30},\\
c^{\mathrm{morn}}_{hdt} &= x_{hdt}-\bar{x}_{hd,02{:}00\text{--}05{:}30},
&t&\in 06{:}00\text{--}09{:}30,\\
c^{\mathrm{mid}}_{hdt} &= x_{hdt}-\bar{x}_{hd,08{:}00\text{--}09{:}30},
&t&\in 10{:}00\text{--}15{:}30.
\end{aligned}
\label{eq:ev-contrasts}
\end{equation}
An evening or overnight signal requires at least one bin meeting or exceeding $\max\{Q^{\mathrm{eve}}_{0.80,h},0.60\ \mathrm{kWh}\}$, where $Q^{\mathrm{eve}}_{0.80,h}$ is the household's 80th percentile across its evening and overnight contrasts. Morning and midday signals require three consecutive bins meeting or exceeding 0.70\,kWh. On PV-active days, the morning gate increases to 0.90\,kWh and the midday branch is disabled. A qualifying day must also reach the household's 40th-percentile daily total. At 30-minute resolution, the 0.60, 0.70, and 0.90\,kWh gates correspond to average-power differences of 1.2, 1.4, and 1.8\,kW. The household quantiles account for habitual variability, the daily-energy floor excludes unusually low-use days, and the consecutive-bin rule suppresses isolated changes. The thresholds serve as consistent rules for identifying charging-like activity; charger measurements were unavailable for calibration or validation. A 21-day window is PV-positive if it contains at least one PV-active day and EV-positive if it contains at least three EV-active days; the EV threshold is raised to four when a PV-active day occurs in the same window to reduce PV--EV confounding. Windows overlapping installation or transition periods are removed with a $\pm 14$-day buffer, and windows are drawn only within contiguous daily blocks so that they never span missing days.

The supervised task is defined at the 21-day window level: the model detects windows showing PV- or EV-associated activity patterns in AMI traces, providing a practical screening tool for DSM targeting and planning applications. PV labels are supported by export-stream availability, and EV labels are based on repeated charging-like import signatures. Because the labels capture recurring technology-related activity in the meter trace, they align well with DSM and feeder planning, where repeated activity is often more actionable than ownership status alone.

For clustering, days with extended gaps or zero total energy are dropped, duplicates are removed, and each day is $\ell_1$-normalized by its own total before per-day standardization, so that clustering reflects the shape of each day rather than its total consumption. For classification, days with at least 44 of 48 intervals are retained, gaps of up to one hour are linearly interpolated \cite{lepot2017}, and longer or pathological gaps are dropped. Each 21-day window is standardized to zero mean and unit variance. Splits are by household to prevent leakage: 20\% of households are held out for testing, then 20\% of the remaining households are used for validation, with all windows from a household assigned to the same split.

\section{Methods}\label{sec:method}
\subsection{Shape-aware daily-profile clustering}
Given daily profiles $\mathcal{S} = \{x_i\}_{i=1}^N$ for a fixed cohort and day type, we partition days into $k$ behaviorally coherent groups using DTW $k$-means. PV-only and co-adoption subsets use export profiles, while EV-only and neither-adopter subsets use import profiles. DTW is used because residential load-shape features such as midday troughs, charging-like peaks, or evening ramps can be shifted in time; Euclidean distance would penalize these shifts even when the underlying shape is similar. Because Euclidean means are not compatible with DTW, cluster prototypes are updated with DBA, which aligns each member to the current prototype and averages in the aligned space \cite{datta2020}. A Sakoe--Chiba band of radius 6 limits the warping path \cite{geler2019}. We use $k=5$ for each of the eight cohort-by-day-type subsets. Five clusters show the main timing patterns without overcrowding each panel, while a fixed value allows the subsets to be compared at the same descriptive resolution. Two random initializations are run per subset, the solution with the lower objective value is retained, and both centroid shapes and cluster sizes are reported.

Formally, for a fixed cohort and day type, the clustering objective is
\begin{equation*}
\min_{\{\mu_j\}_{j=1}^{k},\{c(i)\}_{i=1}^{N}}
\sum_{i=1}^{N} D_{\mathrm{DTW}}\big(x_i,\mu_{c(i)}\big),
\end{equation*}
where $c(i)$ gives the cluster assigned to profile $x_i$ and $\mu_j$ is the DBA centroid for cluster $j$. Because profiles are clustered day by day, the resulting archetypes describe recurring daily load shapes. Different days from the same household may appear in different clusters as their daily import or export patterns change.

\subsection{Multi-label technology detection}
The classifier maps a 21-day window $X \in \mathbb{R}^{T \times F}$ to probabilities $\hat{p} = [\hat{p}_{\mathrm{PV}},\hat{p}_{\mathrm{EV}}]$, with decisions
\[
\hat{y}_{\ell} =  \mathbf{1}\{\hat{p}_{\ell} \ge \tau_{\ell}\},\quad \ell \in \{\mathrm{PV},\mathrm{EV}\}.
\]
The BiLSTM, a sequence encoder that has proved effective on smart-grid metering data \cite{khalid2024}, encodes each 21-day sequence of daily feature vectors: 48 z-scored half-hour import readings, six simple summaries, a calendar day-type indicator, and the PV-activity flag described above ($F = 56$). Because this flag is derived from the export stream, the reported classification is assisted by export information as well as the import readings. Forward and backward final states are concatenated and fed to a two-unit linear head, with sigmoid outputs giving PV and EV probabilities. Training minimizes the sum of two class-weighted binary cross-entropies, with positive-class weights set from training prevalence. Optimization uses adaptive moment estimation with decoupled weight decay (learning rate $10^{-4}$, weight decay $10^{-4}$), dropout 0.6, gradient clipping at 1.0, batch size 64, reduce-on-plateau learning-rate scheduling, and early stopping on validation macro-F1. Results are reported as means over three seeded runs.

The multi-label formulation is used because PV and EV activity can co-occur within the same 21-day window; a four-class softmax would force the model to learn mutually exclusive classes even though the underlying technologies are not mutually exclusive \cite{Bogatinovski2022}. A household-window can be positive for neither, either, or both. Separate sigmoid outputs also allow the PV and EV decision thresholds to be tuned independently. The two labels have different signal characteristics: PV affects many daylight intervals and is supported by export metadata in the current feature set, whereas EV activity can be intermittent and more easily confused with other high-power household loads. Independent thresholds allow the operating point to reflect these different error profiles.

\subsection{Baselines and evaluation}
Logistic regression, random forest, and extreme gradient boosting (XGBoost) baselines use the same 21-day windows summarized into tabular features: mean, standard deviation, minimum, maximum, and selected quantiles (10th, 50th, and 90th percentiles). Logistic regression uses $\ell_2$ regularization with class weights; random forest uses 600 trees with square-root feature subsampling and class balancing; and XGBoost uses 600 estimators, a learning rate of 0.05, a maximum depth of 6, and positive-class weighting. For all models, per-label thresholds $(\tau_{\mathrm{PV}},\tau_{\mathrm{EV}})$ are tuned on the validation set through a grid over $[0.30, 0.70]$ in steps of 0.05 to maximize macro-F1, then applied unchanged to the test set. We report accuracy, macro-precision, macro-recall, macro-F1, and mean area under the receiver operating characteristic curve (AUROC), plus per-class precision, recall, and F1 for the four adoption states. Robustness is assessed through label-sensitivity analysis and strictly forward temporal splits.

Simple and tree-based baselines provide context for the BiLSTM result. An advantage over the tabular summaries would suggest that within-window ordering carries useful information; comparable XGBoost performance would mean a lower-complexity model may suffice for many utility detection workflows. The evaluation therefore emphasizes per-class precision and recall alongside the headline AUROC, since a high AUROC can coexist with an unsuitable deployment threshold, especially for imbalanced or operationally asymmetric labels such as EV-only windows.

On an AMD Ryzen 7 5700X CPU, the eight clustering fits required 44.3 cumulative minutes, and mean classifier fitting times over three seeds were 171.4 seconds for BiLSTM, 19.7 seconds for XGBoost, 23.9 seconds for random forest, and 73.9 seconds for logistic regression. These are offline fitting times excluding preprocessing and inference. Household windows can be scored independently in batches, and deployment throughput should be verified on the intended utility platform.

\section{Results}\label{sec:results}
\subsection{Behavioral archetypes}
Because the profiles are normalized, the centroids show relative timing rather than absolute import or export energy. Cluster sizes are reported in Table~\ref{tab:cluster-sizes}.

\begin{figure*}[tbp]
  \centering
  \includegraphics[width=\textwidth,height=0.93\textheight,keepaspectratio]{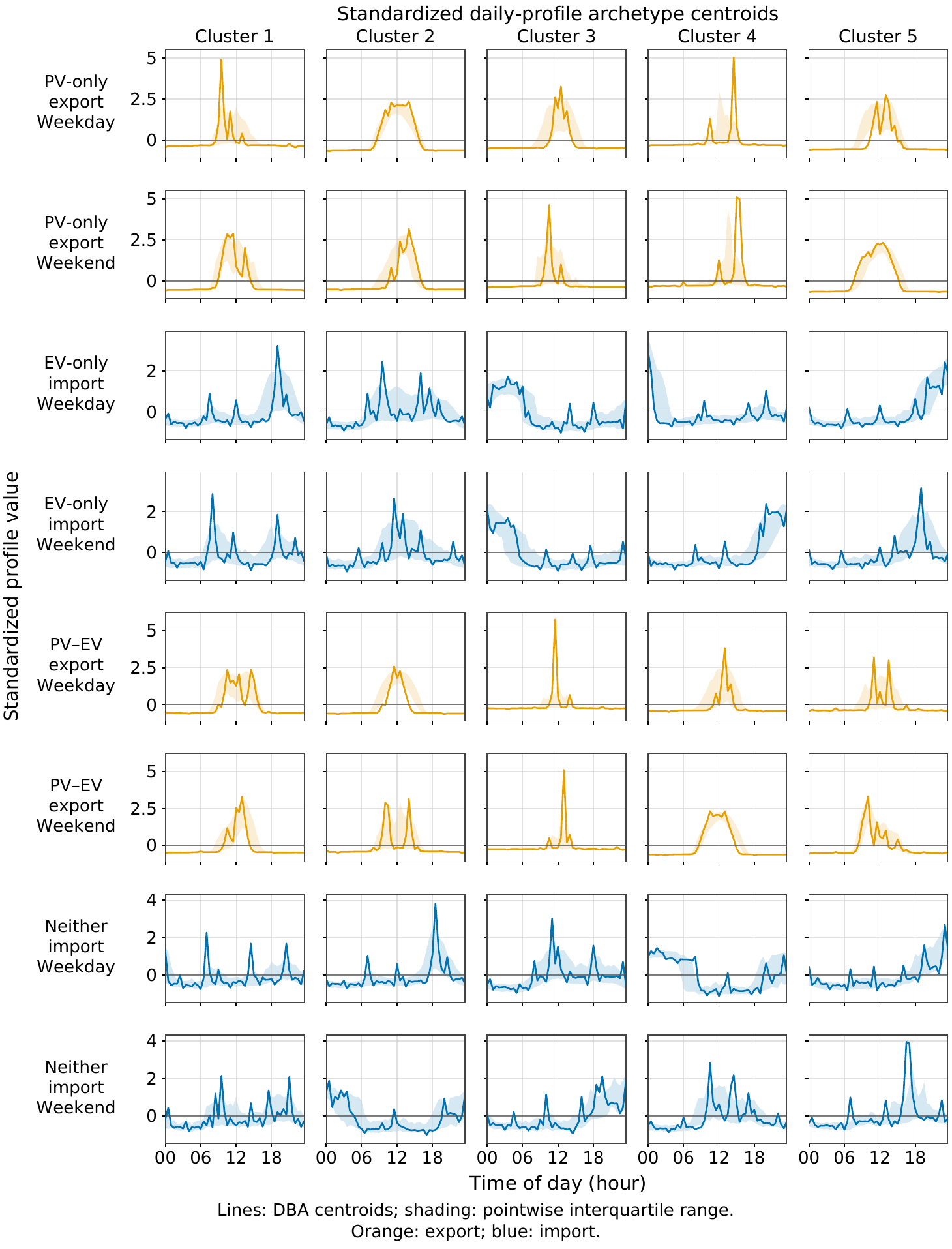}
  \caption{Standardized DTW $k$-means daily-profile archetypes across the four adoption cohorts. Rows identify cohort, clustering channel, and day type; columns identify the panel-local clusters 1--5. Orange lines show export centroids for the PV-associated cohorts, and blue lines show import centroids for the EV-only and neither-adopter cohorts. Shading shows the pointwise interquartile range of the standardized profiles assigned to each cluster.}
  \label{fig:archetype-clusters}
\end{figure*}

Across weekday and weekend PV-only plots, the export centroids show little overnight export, a daytime plateau from late morning to midday, and an afternoon decline (Fig.~\ref{fig:archetype-clusters}, first two rows). Weekdays mix sharp morning or evening-spike clusters with broader daytime shoulders, while weekends show less uniform daytime export with earlier peaking and wider afternoon plateaus. Sizes in Table~\ref{tab:cluster-sizes} show a dominant weekday archetype ($\approx 40\%$; 504/1,267), two sizeable patterns (289 and 274), and two smaller spike clusters (92 and 108). Weekends also have one dominant archetype ($\approx 45\%$; 244/548), two moderate clusters (117 and 107), and two smaller spike clusters (51 and 29). These clusters show substantial within-cohort variation in the timing and concentration of grid export.

EV-only centroids are dominated by charging-like peaks with timing variation. Weekday demand is mostly low during the day and includes overnight plateaus, while morning and late-day spikes may reflect pre- and post-work plug-ins (Fig.~\ref{fig:archetype-clusters}, EV-only rows). Weekends exhibit greater variety, including a midday-spike pattern. Sizes in Table~\ref{tab:cluster-sizes} are fairly even on weekends; on weekdays, morning- and evening-spike profiles are largest (451 in cluster 1 and 479 in cluster 2). Overnight plateaus and weekend midday spikes are consistent with scheduled or opportunistic charging, although aggregate import cannot confirm the responsible end use.

Co-adoption export profiles concentrate from daytime to late afternoon (Fig.~\ref{fig:archetype-clusters}, co-adoption rows). Weekdays include a narrow mid-morning or noon spike and later ramps with softer afternoon and early-evening peaks. Sizes in Table~\ref{tab:cluster-sizes} show a dominant weekday archetype ($\approx 50\%$; 1,654/3,332) with a midday-centered plateau. Weekends are more balanced but still led by a similar plateau ($\approx 46\%$; 662/1,431). The cluster patterns show export concentrated within the solar window, with tight midday centering on weekdays relaxing slightly on weekends. Charger-level validation would be required to relate these export patterns to charging behavior.

Neither-adopter profiles are more varied and spiky, with no consistent midday dip, while still retaining weekday and weekend differences (Fig.~\ref{fig:archetype-clusters}, final two rows). Weekdays show distinct bursts in the morning, at midday, and in the evening, plus an overnight or early-morning plateau that could resemble EV charging. Weekends exhibit smaller spikes spread across the day, including mid-morning and afternoon activity and a clear evening spike. Sizes in Table~\ref{tab:cluster-sizes} are more balanced, although the evening-spike centroid is most prominent (302/880), consistent with heterogeneous loads that carry no clear PV/EV signature.

Across cohorts, PV-only and co-adoption export profiles are organized around solar-window timing, whereas EV-only import profiles contain a wider spread of charging-like peaks. Because these cohorts are clustered on different channels, their standardized centroids do not establish differences in charging timing. The neither cohort shows that not every spike or plateau is technology-specific. Background residential loads can create visually similar shapes, which helps explain why EV-only remains the most difficult class in the detection task.

These patterns show why cohort- and channel-aware interpretation is important. The EV-only import archetypes capture heterogeneous demand timing, while the PV-only and co-adoption export archetypes capture heterogeneous export timing and concentration. A planning workflow should preserve this within-cohort variation and distinguish import from export when linking archetypes to interventions. The cluster sizes indicate which patterns are common enough to matter for program design, and the raw-energy metrics in \ref{app:raw-metrics} quantify their operating scale.

\begin{table}[tbp]
\centering
\caption{Cluster sizes for behavioral archetypes ($k = 5$ for all groups).}
\label{tab:cluster-sizes}
\small
\setlength{\tabcolsep}{4pt}
\begin{tabular}{lcccccc}
\hline
 & \multicolumn{5}{c}{\textbf{Cluster sizes}} & \\
 & \textbf{1} & \textbf{2} & \textbf{3} & \textbf{4} & \textbf{5} & \textbf{Total} \\
\hline
\textbf{PV-only weekday} & 92 & 504 & 289 & 108 & 274 & 1,267 \\
\textbf{PV-only weekend} & 117 & 107 & 51 & 29 & 244 & 548 \\
\textbf{EV-only weekday} & 451 & 479 & 241 & 196 & 299 & 1,666 \\
\textbf{EV-only weekend} & 120 & 160 & 160 & 108 & 165 & 713 \\
\textbf{Co-adoption weekday} & 563 & 1,654 & 254 & 524 & 337 & 3,332 \\
\textbf{Co-adoption weekend} & 310 & 155 & 109 & 662 & 195 & 1,431 \\
\textbf{Neither weekday} & 152 & 302 & 222 & 82 & 122 & 880 \\
\textbf{Neither weekend} & 80 & 55 & 63 & 87 & 90 & 375 \\
\hline
\end{tabular}
\end{table}

Appendix Tables~\ref{tab:raw-weekday} and~\ref{tab:raw-weekend} report day-weighted cluster means for daily import, peak import, evening import and peak, midday import and export, and peak-to-average ratio. Weekday EV-only cluster 5 has the largest evening import and evening peak (14.72\,kWh and 4.18\,kW), while the largest midday exports occur for weekday PV-only cluster 2 and co-adoption cluster 2 (11.21 and 10.71\,kWh) and weekend PV-only cluster 5 and co-adoption cluster 4 (12.35 and 12.97\,kWh).

\subsection{Technology detection performance}

All models achieve high discrimination, with AUROC values ranging from approximately 0.98 to 0.99 (Table~\ref{tab:model_comparison}). The BiLSTM achieves accuracy $= 0.926$, AUROC $= 0.991$, and precision/recall/F1 $= 0.907/0.906/0.906$. XGBoost matches it closely and is marginally higher in precision, recall, and F1 (precision/recall/F1 $= 0.908/0.908/0.908$). Random forest and logistic regression trail by approximately 0.5--2.8\% in F1/AUROC but remain competitive. Per-seed variability is small for all models ($\sigma \in [0.0003,0.0035]$), so these rankings are stable across runs. The close performance of the BiLSTM and XGBoost indicates that the windowing and thresholding protocol is an important part of the detection workflow, while the BiLSTM provides a useful recall-oriented operating point for EV-only windows and XGBoost is marginally stronger on aggregate macro-F1. The EV-only class is difficult because it is more behaviorally heterogeneous and more easily confused with other high-power household loads.

Per-class results (Table~\ref{tab:per-class-metrics}) show that PV-only is the easiest class: the BiLSTM achieves precision/recall/F1 $= 0.981/0.965/0.973$, with XGBoost close behind (F1 $= 0.970$). Co-adoption performance is also strong; the BiLSTM has the highest recall and F1 (recall/F1 $= 0.958/0.939$), ahead of XGBoost (F1 $= 0.930$) and random forest (F1 $= 0.889$). Neither is moderately challenging: the BiLSTM (F1 $= 0.866$) and XGBoost (F1 $= 0.878$) are comparable; random forest trades precision for recall (precision/recall $= 0.776/0.941$), over-flagging neither. EV-only is the most difficult class because it has the smallest sample size and charging-like behavior can be confused with that of household appliances. The BiLSTM gives the highest recall and a strong F1 (recall/F1 $= 0.836/0.832$); XGBoost has slightly lower recall (0.820) but higher precision and F1 (precision/F1 $= 0.861/0.840$). Random forest has lower recall and F1 (recall/F1 $= 0.659/0.760$); logistic regression performs reasonably (F1 $= 0.808$) but remains behind the BiLSTM and XGBoost.

The high AUROC and similar macro-F1 across models highlight the strength of the windowing and thresholding protocol, while the BiLSTM's advantage in EV-only recall shows the value of sequence modeling for the most difficult class. The PV/EV activity rules, together with 21-day windows, a 7-day stride, and transition buffers, produce separable targets, especially for PV-only and co-adoption. Small changes to the labeling rules can shift precision and recall, motivating the sensitivity checks in Section~\ref{sec:robustness}.

Thresholds are tuned on the validation set for macro-F1 under class imbalance. The BiLSTM's consistently high recall for co-adoption and EV-only indicates effective use of weekly repetition and multi-day charging patterns. Logistic regression attains very high co-adoption recall (0.954) with lower precision (0.851). Random forest attains high recall for neither (0.941) but lower precision (0.776), indicating over-prediction of background patterns. Because the aggregate results are close, the per-class error profile matters more than a single headline score. XGBoost is more conservative on EV-only, with higher precision and lower recall, while the BiLSTM gives the opposite trade-off. Higher recall may suit broad DSM outreach; higher EV-only precision may suit targeted tariff offers or feeder studies where false positives could distort planning assumptions.

\begin{table}[tbp]
\centering
\caption{Comparison of model performance (mean over three runs).}
\label{tab:model_comparison}
\small
\resizebox{\columnwidth}{!}{%
\begin{tabular}{@{}l *{5}{S[table-format=1.3]}@{}}
\toprule
\textbf{Model} & \textbf{Accuracy} & \textbf{Precision} & \textbf{Recall} & \textbf{F1} & \textbf{AUROC} \\
\midrule
BiLSTM            & \textbf{0.926} & 0.907          & 0.906          & 0.906          & \textbf{0.991} \\
XGBoost           & 0.926          & \textbf{0.908} & \textbf{0.908} & \textbf{0.908} & 0.990          \\
Random forest     & 0.910          & 0.895          & 0.890          & 0.892          & 0.986          \\
Logistic regression & 0.903       & 0.880          & 0.881          & 0.880          & 0.980          \\
\bottomrule
\end{tabular}
}
\end{table}

\begin{table*}[!tb]
\centering
\caption{Comparison of per-class classification metrics across models.}
\label{tab:per-class-metrics}
\small
\setlength{\tabcolsep}{16pt}
\begin{tabular}{l l *{3}{S[table-format=1.3]} c}
\toprule
\textbf{Model} & \textbf{Class} & \textbf{Precision} & \textbf{Recall} & \textbf{F1 score} & \textbf{Sample size} \\
\midrule
\multirow{4}{*}{BiLSTM sequence}
& PV-only & 0.981 & 0.965 & 0.973 & 4,359 \\
& EV-only & 0.828 & 0.836 & 0.832 & 1,539 \\
& Co-adoption & 0.920 & 0.958 & 0.939 & 1,897 \\
& Neither & 0.871 & 0.862 & 0.866 & 1,935 \\
\midrule
\multirow{4}{*}{XGBoost}
& PV-only & 0.969 & 0.970 & 0.970 & 4,359 \\
& EV-only & 0.861 & 0.820 & 0.840 & 1,539 \\
& Co-adoption & 0.931 & 0.929 & 0.930 & 1,897 \\
& Neither & 0.862 & 0.895 & 0.878 & 1,935 \\
\midrule
\multirow{4}{*}{Random forest}
& PV-only & 0.936 & 0.975 & 0.955 & 4,359 \\
& EV-only & 0.898 & 0.659 & 0.760 & 1,539 \\
& Co-adoption & 0.938 & 0.846 & 0.889 & 1,897 \\
& Neither & 0.776 & 0.941 & 0.850 & 1,935 \\
\midrule
\multirow{4}{*}{Logistic regression}
& PV-only & 0.979 & 0.927 & 0.952 & 4,359 \\
& EV-only & 0.826 & 0.791 & 0.808 & 1,539 \\
& Co-adoption & 0.851 & 0.954 & 0.900 & 1,897 \\
& Neither & 0.839 & 0.868 & 0.853 & 1,935 \\
\bottomrule
\end{tabular}
\end{table*}

\section{Sensitivity and Temporal Generalization}\label{sec:robustness}
This section stress-tests the BiLSTM beyond the baseline comparisons in Section~\ref{sec:results}. Robustness is assessed through sensitivity to plausible changes in the day-level rules used to generate window labels and through temporal generalization, where models are trained on earlier months and tested strictly forward.

Across all variants, performance is stable: macro-F1 $= 0.894$ to $0.914$, AUROC $= 0.988$ to $0.992$, and accuracy $= 0.915$ to $0.934$ on the same 9,730-window test split. The baseline performs well (F1 $= 0.904$, AUROC $= 0.991$, accuracy $= 0.924$), and the additional settings show low sensitivity to changes in the energy floor, absolute contrast gates, and number of EV-active days. Requiring at least four EV-active days per window improves all headline metrics (F1 $= 0.914$, AUROC $= 0.992$, accuracy $= 0.934$), consistent with reduced label noise and better separability. Validation-selected PV thresholds remain low (0.30 to 0.35), whereas EV thresholds are higher (0.55 to 0.70), consistent with the per-class difficulty.

The sensitivity variants cover household evening quantiles from 0.75 to 0.85, daily-total quantiles from 0.35 to 0.45, daytime gates from 0.60 to 0.80\,kWh, evening and overnight gates from 0.50 to 0.70\,kWh, and non-PV window minima from two to four EV-active days.

Stricter EV-positive window rules improve headline scores, consistent with reduced label noise: windows with more repeated charging-like days are easier to distinguish from background residential variability. A strict rule identifies a cleaner set of high-confidence EV-active windows; a looser rule captures more marginal or occasional home-charging behavior at the cost of additional ambiguity. The appropriate rule depends on whether the utility objective is conservative feeder-level detection, broad customer outreach, or fine-grained behavioral analysis.

Forward temporal evaluation remains strong with minor seasonal variation (Table~\ref{tab:temporal-splits}). The mid-year split shows a small dip (F1 $= 0.894$), and performance recovers in the final split (F1 $= 0.906$). Validation-selected thresholds are stable across folds (PV $\approx 0.30$, EV $= 0.50$ to $0.60$), so frequent retuning is unnecessary.

The BiLSTM sustains high discrimination (AUROC $\approx 0.99$) and balanced accuracy/F1 under both rule perturbations and forward splits, with stable operating thresholds across label settings and limited temporal degradation.

PV output, household routines, and EV charging are all seasonal, and the modest mid-year dip indicates that temporal distribution shift is present but mild under the tested splits.

\begin{table*}[!tb]
\centering
\caption{BiLSTM performance under alternative label-generation rules (same test split). Note that other variants include alternative EV-active-day thresholds, energy floors, and contrast gates; full definitions are provided in the text.}
\label{tab:parameter-settings}
\small
\setlength{\tabcolsep}{10pt}
\begin{tabular}{l *{3}{S[table-format=1.3]} *{2}{S[table-format=1.2]}}
\toprule
\textbf{Setting (label rule)} & \textbf{F1} & \textbf{AUROC} & \textbf{Accuracy} & \textbf{PV threshold} & \textbf{EV threshold} \\
\midrule
Baseline                          & 0.904 & 0.991 & 0.924 & 0.30 & 0.70 \\
EV window: $\geq 4$ active days   & 0.914 & 0.992 & 0.934 & 0.30 & 0.65 \\
Higher contrast gate              & 0.911 & 0.992 & 0.931 & 0.30 & 0.55 \\
Looser energy floor (35th percentile) & 0.909 & 0.991 & 0.927 & 0.35 & 0.55 \\
Other variants (range)            & {0.894--0.905} & {0.988--0.991} & {0.915--0.925} & {0.30--0.70} & {0.55--0.70} \\
\bottomrule
\end{tabular}
\end{table*}

\begin{table}[tbp]
\centering
\caption{BiLSTM model performance across temporal data splits.}
\label{tab:temporal-splits}
\small
\resizebox{\columnwidth}{!}{%
\begin{tabular}{@{}l *{3}{S[table-format=1.3]} *{2}{S[table-format=1.2]}@{}}
\toprule
\tabincell{c}{\textbf{Training/}\\\textbf{validation cutoff}} & \textbf{F1} & \textbf{AUROC} & \textbf{Accuracy} & {\tabincell{c}{\textbf{PV}\\\textbf{threshold}}} & {\tabincell{c}{\textbf{EV}\\\textbf{threshold}}} \\
\midrule
31 Mar / 30 Jun & 0.898 & 0.990 & 0.920 & 0.30 & 0.50 \\
30 Jun / 30 Sep & 0.894 & 0.989 & 0.914 & 0.30 & 0.60 \\
30 Sep / 31 Dec & 0.906 & 0.991 & 0.930 & 0.30 & 0.50 \\
\bottomrule
\end{tabular}
}
\end{table}

\section{Discussion}\label{sec:discussion}
This study investigated residential PV/EV behaviors and PV/EV-associated activity detection from aggregate smart-meter data using a two-part methodology. In the first part, DTW $k$-means with DBA centroids was used to produce daily load-shape clusters by cohort and day type. PV-only export profiles consistently showed midday concentration, while EV-only import profiles exhibited charging-like peaks at distinct times, including overnight plateaus and pre- and post-work spikes. Co-adoption export profiles were more concentrated around midday on weekdays, and households in the neither category displayed heterogeneous, spikier import profiles. In the second part, a BiLSTM trained on 21-day windows with activity labeling and validation-tuned thresholds achieved high discrimination and a competitive macro-F1 relative to strong tabular baselines. Given their similar aggregate performance, XGBoost is a practical choice for simpler deployment and for applications that value its higher EV-only precision. The BiLSTM is better suited to recall-oriented screening where missing EV-active windows is more costly.

Shape-aware clustering provides interpretable grid-exchange patterns that can inform DSM targeting. The classification analysis shows that sequence models can detect PV/EV activity at scale under the labeling protocol used here, supporting targeted tariffs, customer outreach, and feeder-level planning. Label and window design strongly influence outcomes, so precise activity rules and decision thresholds are an important part of the detection workflow. Combining the two analyses allows the clustering to give operational meaning to the labels and the classification to provide a scalable route for identifying relevant households or windows.

These outputs can help utilities prioritize where interventions are plausible. Co-adoption archetypes with large midday export may be candidates for programs that encourage flexible load during the solar window, while EV-only import archetypes with overnight or evening peaks may warrant further managed-charging assessment or tariff designs that avoid synchronized rebound peaks. These archetypes describe observed load shapes and should be considered alongside program design, customer engagement, and network constraints when inferring customer behavior. Practical screening can use evening energy and peak demand to identify managed-charging candidates; midday import and export to assess solar-following flexible-load offers and reverse-flow concerns; and peak-to-average ratio to flag synchronization or local loading concerns, with device-level confirmation where required.

In low-voltage network planning, demand timing determines whether PV and EV activities contributes to network constraints. The archetype results describe differences in import and export timing across the adoption cohorts. Although feeder-level power-flow analysis is beyond the scope of this study, the detected activity windows and archetype memberships provide behavioral inputs for future feeder-level coincidence and loading studies, including estimates of evening peak demand, midday reverse power flow, ramping risk associated with each archetype, and DER coordination \cite{yang2021exploring}.

Several limitations should be considered. The normalization used for clustering emphasizes timing and shape, so the archetypes do not represent exact schedules. PV-associated cohorts are clustered on export and the other cohorts on import, which limits direct cross-cohort inference about charging timing. The EV labels may also capture high-power loads such as pool or spa heating, water heating, or space conditioning, and charger data were not available for validation. PV association depends on observable export and may omit export-limited or zero-export systems. The classifier is assisted by the export-derived PV-activity flag. De-identified AMI traces require appropriate privacy and data-governance controls. The household hold-out and temporal splits are based on AusNet data from Victoria, Australia; applying the workflow to other climates, tariffs, customer behaviors, metering practices, or distribution networks will require local validation and recalibration.

Future work should focus on three priorities:
\begin{itemize}
    \item Incorporate weather and export-channel information to test whether richer feature sets further improve PV and EV detection.
    \item Improve the classifier and labeling rules by comparing the current BiLSTM with other sequence baselines and validating the window labels against household EV ownership metadata, charger records, or rebate timing where available.
    \item Evaluate the operational value of the archetype metrics in managed-charging, solar-following flexible-load, and tailored-tariff programs.
\end{itemize}

\section{Conclusion}\label{sec:conclusion}
We have presented a scalable, interpretable pipeline that pairs shape-aware clustering with sequence-based detection to characterize household grid-exchange patterns and detect PV/EV activity from aggregate smart-meter data. The clustering reveals interpretable co-adoption export patterns, most notably a dominant midday-centered weekday archetype, and the BiLSTM detects PV/EV-associated activity windows with high discrimination and the strongest recall on the most difficult EV-only class. The EV-active windows are inferred from load profiles because charger records were not available. The workflow shows promise for demand-side planning and will require stronger technology-level ground truth and external validation before full operational deployment.

\appendix
\section{Raw-Energy Metrics for the Archetypes}\label{app:raw-metrics}
Daily import is the sum of the 48 half-hour readings; peak import is twice the largest half-hour reading. Evening metrics cover 18:00--23:30 and midday metrics cover 10:00--15:30. Peak-to-average ratio is the maximum import divided by mean half-hour import. The tables report day-weighted arithmetic means for each cluster; export means for PV-associated cohorts are conditional on the positive-export days retained for clustering.

\begin{table*}[!b]
\centering
\caption{Weekday raw-energy means by daily archetype. A dash denotes a cohort without an export stream.}
\label{tab:raw-weekday}
\footnotesize
\setlength{\tabcolsep}{3.2pt}
\renewcommand{\arraystretch}{1.05}
\begin{tabular*}{\textwidth}{@{\extracolsep{\fill}}llrrrrrrr@{}}
\toprule
\textbf{Cohort} & \textbf{Cluster} &
\tabincell{c}{\textbf{Daily import}\\\textbf{(kWh)}} &
\tabincell{c}{\textbf{Peak import}\\\textbf{(kW)}} &
\tabincell{c}{\textbf{Evening import}\\\textbf{(kWh)}} &
\tabincell{c}{\textbf{Evening peak}\\\textbf{(kW)}} &
\tabincell{c}{\textbf{Midday import}\\\textbf{(kWh)}} &
\tabincell{c}{\textbf{Midday export}\\\textbf{(kWh)}} &
\tabincell{c}{\textbf{Peak-to-}\\\textbf{average ratio}} \\
\midrule
PV-only & 1 & 15.87 & 2.75 & 6.02 & 2.01 & 2.89 & 1.23 & 5.11 \\
PV-only & 2 & 9.40 & 1.97 & 4.16 & 1.54 & 0.26 & 11.21 & 5.73 \\
PV-only & 3 & 12.41 & 2.33 & 4.33 & 1.65 & 1.38 & 4.95 & 5.61 \\
PV-only & 4 & 15.89 & 2.65 & 4.96 & 1.67 & 2.84 & 0.64 & 5.15 \\
PV-only & 5 & 12.10 & 2.66 & 4.86 & 1.87 & 1.20 & 5.74 & 5.69 \\
\midrule
EV-only & 1 & 18.90 & 3.31 & 8.05 & 2.91 & 3.33 & -- & 4.45 \\
EV-only & 2 & 18.69 & 2.68 & 4.35 & 1.42 & 6.64 & -- & 3.80 \\
EV-only & 3 & 28.79 & 3.28 & 5.14 & 1.65 & 3.53 & -- & 2.81 \\
EV-only & 4 & 22.62 & 4.43 & 5.35 & 1.87 & 3.59 & -- & 5.52 \\
EV-only & 5 & 25.77 & 4.22 & 14.72 & 4.18 & 3.91 & -- & 4.28 \\
\midrule
Co-adoption & 1 & 14.86 & 3.09 & 5.58 & 2.05 & 1.33 & 5.77 & 5.39 \\
Co-adoption & 2 & 13.97 & 2.96 & 5.26 & 1.96 & 0.63 & 10.71 & 5.48 \\
Co-adoption & 3 & 25.83 & 3.98 & 8.01 & 2.70 & 5.02 & 0.44 & 4.42 \\
Co-adoption & 4 & 19.67 & 3.74 & 6.58 & 2.48 & 2.76 & 3.55 & 5.17 \\
Co-adoption & 5 & 19.46 & 3.46 & 6.65 & 2.37 & 3.44 & 1.70 & 4.97 \\
\midrule
Neither & 1 & 12.92 & 2.25 & 3.88 & 1.50 & 2.74 & -- & 4.71 \\
Neither & 2 & 11.82 & 2.18 & 4.74 & 1.95 & 2.17 & -- & 5.06 \\
Neither & 3 & 11.30 & 1.58 & 2.90 & 0.94 & 4.03 & -- & 4.37 \\
Neither & 4 & 23.92 & 2.31 & 5.34 & 1.68 & 3.21 & -- & 2.38 \\
Neither & 5 & 15.56 & 2.41 & 7.46 & 2.34 & 2.65 & -- & 4.65 \\
\bottomrule
\end{tabular*}
\end{table*}

\begin{table*}[!th]
\centering
\caption{Weekend raw-energy means by daily archetype. A dash denotes a cohort without an export stream.}
\label{tab:raw-weekend}
\footnotesize
\setlength{\tabcolsep}{3.2pt}
\renewcommand{\arraystretch}{1.05}
\begin{tabular*}{\textwidth}{@{\extracolsep{\fill}}llrrrrrrr@{}}
\toprule
\textbf{Cohort} & \textbf{Cluster} &
\tabincell{c}{\textbf{Daily import}\\\textbf{(kWh)}} &
\tabincell{c}{\textbf{Peak import}\\\textbf{(kW)}} &
\tabincell{c}{\textbf{Evening import}\\\textbf{(kWh)}} &
\tabincell{c}{\textbf{Evening peak}\\\textbf{(kW)}} &
\tabincell{c}{\textbf{Midday import}\\\textbf{(kWh)}} &
\tabincell{c}{\textbf{Midday export}\\\textbf{(kWh)}} &
\tabincell{c}{\textbf{Peak-to-}\\\textbf{average ratio}} \\
\midrule
PV-only & 1 & 12.25 & 2.54 & 4.81 & 1.73 & 1.50 & 5.72 & 5.41 \\
PV-only & 2 & 13.90 & 2.60 & 5.01 & 1.70 & 1.48 & 6.12 & 6.04 \\
PV-only & 3 & 24.87 & 3.93 & 7.80 & 2.63 & 5.77 & 1.55 & 3.91 \\
PV-only & 4 & 15.97 & 2.75 & 4.86 & 1.46 & 2.84 & 0.77 & 6.67 \\
PV-only & 5 & 8.15 & 1.65 & 3.67 & 1.33 & 0.23 & 12.35 & 5.44 \\
\midrule
EV-only & 1 & 16.26 & 2.62 & 4.64 & 1.57 & 3.64 & -- & 4.50 \\
EV-only & 2 & 20.39 & 3.19 & 3.94 & 1.31 & 8.97 & -- & 4.07 \\
EV-only & 3 & 28.26 & 3.95 & 5.16 & 1.75 & 3.64 & -- & 3.79 \\
EV-only & 4 & 23.25 & 3.64 & 12.80 & 3.61 & 4.00 & -- & 4.15 \\
EV-only & 5 & 23.36 & 4.03 & 9.70 & 3.46 & 4.76 & -- & 4.43 \\
\midrule
Co-adoption & 1 & 18.64 & 3.42 & 6.14 & 2.08 & 2.48 & 5.72 & 5.17 \\
Co-adoption & 2 & 20.22 & 3.66 & 6.40 & 2.25 & 3.76 & 2.95 & 4.77 \\
Co-adoption & 3 & 27.20 & 4.37 & 7.68 & 2.80 & 6.55 & 0.65 & 4.93 \\
Co-adoption & 4 & 13.58 & 2.78 & 5.54 & 1.98 & 0.46 & 12.97 & 5.25 \\
Co-adoption & 5 & 15.77 & 3.17 & 5.58 & 1.94 & 2.72 & 4.96 & 5.19 \\
\midrule
Neither & 1 & 10.26 & 1.55 & 3.17 & 1.23 & 2.56 & -- & 3.65 \\
Neither & 2 & 17.19 & 2.12 & 4.62 & 1.54 & 2.70 & -- & 3.45 \\
Neither & 3 & 16.52 & 2.72 & 7.87 & 2.54 & 2.73 & -- & 4.03 \\
Neither & 4 & 12.36 & 1.94 & 2.69 & 0.96 & 4.90 & -- & 4.07 \\
Neither & 5 & 12.09 & 2.41 & 3.78 & 1.61 & 3.32 & -- & 5.18 \\
\bottomrule
\end{tabular*}
\end{table*}

\section*{Acknowledgments}
This work was supported in part by the Australian Research Council Discovery Early Career Researcher Award under Grant DE230100046.




\bibliographystyle{elsarticle-num} 
\bibliography{ref}
\end{document}